\def\year{2022}\relax
\documentclass[letterpaper]{article} 
\usepackage{aaai22}  
\usepackage{times}  
\usepackage{helvet}  
\usepackage{courier}  
\usepackage[hyphens]{url}  
\usepackage{graphicx} 
\usepackage{booktabs}
\usepackage{multirow}
\usepackage{lipsum}
\usepackage{threeparttable}
\usepackage{amsmath}
\usepackage{amssymb}
\usepackage{bm}
\usepackage{kotex}
\usepackage{color}
\usepackage{graphics}
\usepackage{graphicx}
\usepackage{pifont}
\usepackage{cite}
\usepackage{tabularx}
\usepackage{float}
\usepackage{graphicx}
\usepackage{capt-of}
\DeclareMathAlphabet\mathbfcal{OMS}{cmsy}{b}{n}
\usepackage{amsthm}
\usepackage{arydshln}
\usepackage{adjustbox}

\usepackage{mathtools}

\newcommand{\window}{\bm{w}_{t}}

\newcommand{\anoscore}{\mathcal{A}\left(\window\right)}

\newcommand{\rep}[1]{#1$^{\dagger}$}
\usepackage{relsize,amsfonts}
\newcommand\bigexists{%
  \mathop{\lower0.75ex\hbox{\ensuremath{%
    {{\mathlarger{\mathlarger{\exists}}}}}}}%
  \limits}

\newcommand{\hi}{(\scalebox{0.8}{\textcolor{green}{$\uparrow$}})}
\newcommand{\lo}{(\scalebox{0.8}{\textcolor{red}{$\downarrow$}})}

\usepackage[symbol]{footmisc}

\usepackage{natbib}  
\usepackage{caption} 
\DeclareCaptionStyle{ruled}{labelfont=normalfont,labelsep=colon,strut=off} 
\usepackage{algorithm}
\usepackage{algorithmic}

\usepackage{newfloat}
\usepackage{listings}
\floatstyle{ruled}
\newfloat{listing}{tb}{lst}{}
\floatname{listing}{Listing}
\title{Towards a Rigorous Evaluation of Time-series Anomaly Detection}
\author{
    Siwon Kim\textsuperscript{\rm 1} \quad
    Kukjin Choi\textsuperscript{\rm 1,2} \quad
    Hyun-Soo Choi\textsuperscript{\rm 3,4} \quad
    Byunghan Lee\textsuperscript{\rm 5}\equalcontrib \quad
    Sungroh Yoon\textsuperscript{\rm 1,6}\equalcontrib 
}
\affiliations{
    \textsuperscript{\rm 1} Data Science and AI Laboratory, Seoul National University, Korea, \\
    \textsuperscript{\rm 2} DIT Center, Samsung Electronics, Korea,
    \textsuperscript{\rm 3} Ziovision, Korea,\\
    \textsuperscript{\rm 4} Department of CSE and Education Research Team for Medical Big-data Convergence, Kangwon National University, Korea, \\
    \textsuperscript{\rm 5} Department  of  Electronic  and  IT  Media  Engineering, Seoul National University of Science and Technology, Korea,\\
    \textsuperscript{\rm 6} Department of ECE and Interdisciplinary Program in AI, Seoul National University, Korea,\\
    \texttt{tuslkkk@snu.ac.kr}\qquad
    \texttt{kj21.choi@snu.ac.kr}\qquad 
    \texttt{choi.hyunsoo@kangwon.ac.kr}\\
    \texttt{bhlee@seoultech.ac.kr}\qquad\texttt{sryoon@snu.ac.kr} 
}

\begin{document}
\maketitle

\begin{abstract}
In recent years, proposed studies on time-series anomaly detection (TAD) report high F1 scores on benchmark TAD datasets, giving the impression of clear improvements in TAD. 
However, most studies apply a peculiar evaluation protocol called point adjustment (\texttt{PA}) before scoring. 
In this paper, we theoretically and experimentally reveal that the \texttt{PA}~protocol has a great possibility of overestimating the detection performance; that is, even a random anomaly score can easily turn into a state-of-the-art TAD method.
Therefore, the comparison of TAD methods after applying the \texttt{PA}~protocol can lead to misguided rankings.
Furthermore, we question the potential of existing TAD methods by showing that an untrained model obtains comparable detection performance to the existing methods even when \texttt{PA} is forbidden.
Based on our findings, we propose a new baseline and an evaluation protocol.
We expect that our study will help a rigorous evaluation of TAD and lead to further improvement in future researches.

\end{abstract}

\section{Introduction}\label{section:introduction}
As Industry 4.0 accelerates system automation, consequences of system failures can have a significant social impact~\cite{lee2008cyber, lee2015cyber, baheti2011cyber}.
To prevent this failure, the detection of the anomalous state of a system is more important than ever, and it is being studied under the name of anomaly detection (AD).
Meanwhile, deep learning has shown its effectiveness in modeling multivariate time-series data collected from numerous sensors and actuators of large systems~\cite{chalapathy2019deep}.
Therefore, various time-series AD (TAD) methods have widely adopted deep learning, and each of them demonstrated its own superiority by reporting higher F1 scores than the preceding methods~\cite{choi2021deep}.
For some datasets, the reported F1 scores exceed 0.9, giving an encouraging impression of today's TAD capabilities.  

\begin{figure}[t!]
    \centering
    \includegraphics[width=\columnwidth]{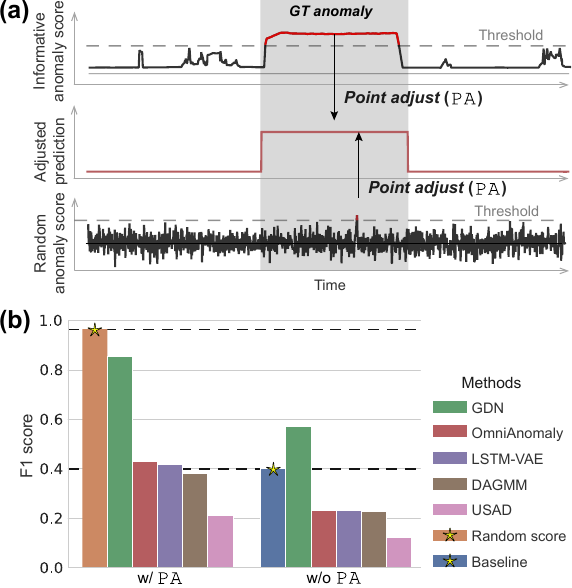}
    \caption{
    (a) \texttt{PA}~makes different anomaly scores indistinguishable. The black lines, gray area, and dashed line indicate the anomaly scores, GT anomaly segment, and TAD threshold, respectively. 
    After applying \texttt{PA}, the predictions for informative and random anomaly scores degenerate to the same adjusted prediction (red). 
    (b) Existing methods fail to exceed $\texttt{F1}_\texttt{PA}$~of a randomly generated anomaly score (left) and show no improvement against the newly proposed baseline (right) even when \texttt{PA}~is forbidden for the WADI dataset. 
    }
    \label{fig:introduction}
\end{figure} 
However, most of the current TAD methods measure the F1 score after applying a peculiar evaluation protocol named point adjustment (\texttt{PA}), proposed by \citeauthor{Xu_2018}~\cite{su2019robust, audibert2020usad, shen2020timeseries}.
\texttt{PA}~works as follows: if at least one moment in a contiguous anomaly segment is detected as an anomaly, the entire segment is then considered to be correctly predicted as anomaly.
Typically, F1 score is calculated with the adjusted predictions (hereinafter denoted by $\texttt{F1}_\texttt{PA}$).
If the F1 score is computed without \texttt{PA}, it is denoted as \texttt{F1}.
The \texttt{PA}~protocol was proposed on the basis that a single alert within an anomaly period is sufficient to take action for system recovery.
It has become a fundamental step in TAD evaluation, and some of the following studies reported only $\texttt{F1}_\texttt{PA}$~without \texttt{F1}~\cite{chen2021learning}. 
A higher $\texttt{F1}_\texttt{PA}$~has been indicated better detection capability. 

However, \texttt{PA}~has a high possibility of overestimating the model performance.
A typical TAD model produces an anomaly score that informs about the degree of input abnormality, and predicts an anomaly if this score is higher than a threshold.
With \texttt{PA}, the prediction from the randomly generated anomaly score and that of the well-trained model becomes identical, as depicted in \figurename~\ref{fig:introduction}-(a).
The black solid lines show two different anomaly scores; the upper line shows informative scores from a well-trained model, while the lower is randomly generated.
The shaded area and dashed line indicate the ground truth (GT) anomaly segment and TAD threshold, respectively.
The informative scores (above) are ideal given that they are high only during the GT segment. 
In contrast, randomly generated anomaly scores (below) cross the threshold only once within the GT segment.
Despite their disparity, the predictions after \texttt{PA}~ become indistinguishable, as indicated by the red line.
If random anomaly scores can yield $\texttt{F1}_\texttt{PA}$~as high as a proficient detection model, it is difficult to conclude that a model with a higher $\texttt{F1}_\texttt{PA}$~performs better than the others.
Our experimental results in Section~\ref{section:results} show that random anomaly scores can overturn most state-of-the-art TAD methods (\figurename~\ref{fig:introduction}-(b)).

Another question that arises is whether \texttt{PA}~is the only problem in the evaluation of TAD methods. 
Until now, only the absolute \texttt{F1}~has been reported, without any attempt to establish a baseline and relative comparison against it. 
If the accuracy of a binary classifier is 50\%, it is not much different from random guessing despite being an apparently large number.
Similarly, a proper baseline should be discussed for TAD, and future methods should be evaluated based on the improvement compared to the baseline.
According to our observations, existing TAD methods do not seem to have obtained a significant improvement over the baseline that this paper proposes. 
Furthermore, several methods fail to exceed it.
Our observations for one of the benchmark dataset are summarized in the right of \figurename~\ref{fig:introduction}-(b).

In this paper, we raise the question of whether the current TAD methods that claim to bring significant improvements are being properly evaluated, and suggest directions for the rigorous evaluation of TAD for the first time. 
Our contributions are summarized as follows:
\begin{itemize}
    \item We show that \texttt{PA}, a peculiar evaluation protocol for TAD, greatly overestimates the detection performance of existing methods.  
    \item We show that, without \texttt{PA}, existing methods exhibit no (or mostly insignificant) improvement over the baseline.
    \item Based on our findings, we propose a new baseline and an evaluation protocol for rigorous evaluation of TAD.
\end{itemize}

\section{Background}\label{section:related_works}


\subsection{Types of anomaly in time-series signals}
Various types of anomalies exist in TAD dataset~\cite{choi2021deep}.
A \textit{contextual anomaly} represents a signal that has a different shape from that of the normal signal.
A \textit{collective anomaly} indicates a small amount of noise accumulated over a period of time.
The \textit{point anomaly} indicates a temporary and significant deviation from the expected range owing to a rapid increase or decrease in the signal value. 
Point anomaly is the most dominant type in the current TAD datasets.

\subsection{Unsupervised TAD}
A typical AD setting assumes that only normal data are accessible during the training time.
Therefore, an unsupervised method is one of the most appropriate approaches for TAD, which trains a model to learn shared patterns in only normal signals.
The final objective is to assign different anomaly scores to inputs depending on the degree of their abnormality, i.e., low and high anomaly scores for normal and abnormal inputs, respectively.
Reconstruction-based AD method trains a model to minimize the distance between a normal input and its reconstruction.
Anomalous input at the test time results in large distance as it is difficult to reconstruct.
The distance, or reconstruction error, serves as an anomaly score.
Forecasting-based approach trains a model to predict the signal that will come after the normal input, and takes the distance between the ground truth and predicted signal as an anomaly score. 
Please refer to Appendix for detailed examples of each category.

\subsection{Assessment of TAD evaluation}
There have been several approaches to point out the pitfalls in current TAD evaluation.
\cite{wu2021current} proposed limitations of the benchmark TAD datasets and shows that a simple detector, so-called one-liner, is sufficient for some datasets.
They also provided several synthetic datasets.
\cite{lai2021revisiting} built a new taxonomy for anomaly types (e.g. point vs. pattern) and introduced new datasets synthesized under new criteria. 
In contrast, we propose the pitfalls in TAD evaluation: the risk of \texttt{PA}'s overestimation and the absence of baseline and the solutions.
If the pitfalls are not resolved, it is impossible to evaluate whether the improvement of a TAD methods is significant even with the better datasets proposed by above papers.

\section{Pitfalls of the TAD evaluation}\label{section:method}


\subsection{Problem formulation}
First, we denote the time-series signal observed from $N$ sensors during time $T$ as $\bm{X}=\{\bm{x}_1, ..., \bm{x}_T\}, \bm{x}_t \in \mathbb{R}^N$.
As conventional approaches, it is normalized and split into a series of windows $\bm{W}=\{\bm{w}_{1}, ..., \bm{w}_{T-\tau+1}\}$ with stride 1, where $\bm{w}_{t}=\{\bm{x}_t, ..., \bm{x}_{t+\tau-1}\}$ and $\tau$ is the window size. 
The ground truth binary label $y_t \in \{0, 1\}$, indicating whether a signal is an anomaly (1) or not (0), is given only for the test dataset.
The goal of TAD is to predict the anomaly label $\hat{y}_t$ for all windows in the test dataset. 
The labels are obtained by comparing anomaly scores $\anoscore$ with a TAD threshold $\delta$ given as follows:
\begin{equation}\label{eq:labeling}
    \hat{y}_{t} = \begin{cases}
    1, &\text{if}\; \anoscore > \delta \\
    0, &\text{otherwise}.
    \end{cases}
\end{equation}

An example of $\anoscore$ is the mean squared error (\texttt{MSE}) between the original input and its reconstructed version, which is defined as follows:
\begin{equation}
\begin{gathered}
    \anoscore = \texttt{MSE}\left(\bm{w}_{t}, \hat{\bm{w}}_{t}\right) =\frac{1}{\tau}\left \| \bm{w}_{t}-\hat{\bm{w}}_{t} \right \|_2
\end{gathered}
\end{equation}\label{eq:anoscore}\noindent{where} $\hat{\bm{w}}_{t} = f_{\theta}\left(\window\right)$ denotes the output from a reconstruction model $f_{\theta}$ parameterized with $\theta$.
After labeling, the precision (P), recall (R), and F1 score for the evaluation are computed as follows:
\begin{equation}\label{eq:metric}
\begin{aligned}  \mathrm{P} = \frac{\mathrm{TP}}{\mathrm{TP}+\mathrm{FP}},&\;\;  \mathrm{R} = \frac{\mathrm{TP}}{\mathrm{TP}+\mathrm{FN}}\\[6pt]
    \mathrm{F1\; score} = &\frac{2 \cdot \mathrm{P} \cdot \mathrm{R}}{\mathrm{P}+\mathrm{R}},
\end{aligned}
\end{equation}
\noindent{where} $\text{TP}$, $\text{FP}$, and $\text{FN}$ denote the number of true positives, false positives and false negatives, respectively.

The TAD test dataset may contain multiple anomaly segments lasting over a few time steps.
We denote $\bm{S}$ as a set of $M$ anomaly segments; that is, $\bm{S}=\{S_1, ...,S_M\}$, where $S_m=\{t^{m}_{s}, ..., t^{m}_{e}\}$; $t^{m}_{s}$ and $t^{m}_{e}$ denote the start and end times of $S_m$, respectively. 
\texttt{PA}~adjusts $\hat{y}_t$ to 1 for all $t \in S_m$ if anomaly score is higher than $\delta$ at least once in $S_m$.
With \texttt{PA}, the labeling scheme of Eq.~\ref{eq:labeling} changes as follows:
\begin{equation}\label{eq:palabeling}
    \hat{y}_{t} = \begin{cases}
    \multirow{2}{*}{1,} & \text{if} \; \anoscore > \delta\\
                        & \text{or}\; t\in S_m \; \text{and} \mathop{\exists}\limits_{t' \in S_m}  \mathcal{A}(\bm{w}_{t'}) > \delta \\[2pt]
    0, &\text{otherwise} .
    \end{cases}
\end{equation}
\noindent{$\texttt{F1}_\texttt{PA}$~denotes} the F1 score computed with adjusted labels.



\subsection{Random anomaly score with high $\texttt{F1}_\texttt{PA}$}\label{sec:pa}
In this section, we demonstrate that the \texttt{PA}~protocol overestimates the detection capability. 
We start from the abstract analysis of the $\mathrm{P}$ and $\mathrm{R}$ of Eq.~\ref{eq:metric}, and we mathematically show that a randomly generated $\anoscore$ can achieve a high $\texttt{F1}_\texttt{PA}$~value close to 1.
According to Eq.~\ref{eq:metric}, as the F1 score is a harmonic mean of $\mathrm{P}$ and $\mathrm{R}$, it also depends on $\mathrm{TP, FN}$, and $\mathrm{FP}$.
As shown in Eq.~\ref{eq:palabeling}, \texttt{PA}~increases $\mathrm{TP}$ and decreases $\mathrm{FN}$ while maintaining $\mathrm{FP}$.
Therefore, after the \texttt{PA}, the P, R and consequently F1 score can only increase. 

\begin{figure}[t!]
    \centering
    \includegraphics[width=0.9\columnwidth]{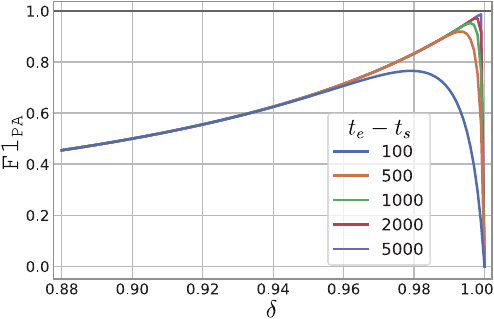}
    \caption{
    $\texttt{F1}_\texttt{PA}$~for the case of uniform random anomaly scores varying with $\delta$ for different $t_e-t_s$.
    If an anomaly segment is considerably long, that is, if $t_e-t_s$ is sufficiently large, $\texttt{F1}_\texttt{PA}$~approaches 1 as $\delta$ increases.
    }
    \label{fig:f1score}
\end{figure}

Next, we show that $\texttt{F1}_\texttt{PA}$~can easily get close to 1.
First, $\mathrm{R}$ is restated as a conditional probability as follows:
\begin{equation}\label{eq:cond}
\begin{aligned}
    \mathrm{R} &= \mathbf{Pr}\left( \hat{y}_t = 1 \mid y_t = 1 \right) \\[5pt]
              &= \mathbf{Pr }\left(\hat{y}_t = 1 \mid t \in \bm{S} \right) \\[5pt]
              &= 1-\mathbf{Pr}\left(\hat{y}_t =0\mid t \in \bm{S}\right). \\
\end{aligned}
\end{equation}
Let's assume that $\anoscore$ is drawn from a uniform distribution $\mathcal{U}(0, 1)$.
We use $0 \leq \delta' \leq 1$ to denote a TAD threshold for this assumption.
If only one anomaly segment exists, i.e., $\bm{S}=\{\{t_s,...,t_e\}\}$, $\mathrm{R}$ after \texttt{PA}~can be expressed as follows, referring to Eq.~\ref{eq:palabeling}:
\begin{equation}\label{eq:recall2}
\begin{aligned}
                \mathrm{R} &= 1 - \mathbf{Pr}\left(\forall \; t' \in \bm{S},~\mathcal{A}(\bm{w}_{t'}) < \delta' \mid t \in \bm{S} \right)\\[5pt]
                &= 1-\prod_{t'\; \in \;\bm{S}} \mathbf{Pr}\left( \mathcal{A}\left( \bm{w}_{t'} \right) < \delta' \mid t \in \bm{S} \right) \\[5pt]
                &=   1 -\frac{1}{\gamma} \cdot \prod_{t'\; \in \;\bm{S}} \mathbf{Pr}\left( \mathcal{A}\left( \bm{w}_{t'} \right) < \delta'\right) \\[5pt]
                &=  1-\frac{1}{\gamma} \cdot\delta'^{\;(t_e-t_s)},
\end{aligned}
\end{equation}
where $\gamma=\mathbf{Pr}(t\in\bm{S})$ is a test dataset anomaly ratio and $\mathbf{Pr}\left(\mathcal{A}\left( \bm{w}_{t'}\right)   < \delta' \right) = \int_0^{\delta'} 1 = \delta'$.

\begin{equation}\label{eq:prec}
\begin{aligned}
    \mathrm{P} &= \mathbf{Pr}\left(y_t = 1 \mid \hat{y}_t=1\right) = \mathrm{R} \cdot \frac{\mathbf{Pr}\left(y_t=1\right)}{\mathbf{Pr}\left(\hat{y}_t = 1\right)}\\[5pt]
               &= \mathrm{R} \cdot \frac{\gamma}{\mathbf{Pr}\left(\hat{y}_t = 1, y_t = 1\right)+\mathbf{Pr}\left(\hat{y}_t = 1, y_t = 0\right)}\\[5pt]
               &= \frac{\gamma-\delta'^{\left(t_e-t_s\right)}}{(\gamma-\delta'^{(t_e-t_s)}) + (1 - \delta')}.\\
\end{aligned}
\end{equation}
For a more generalized proof, please refer to the Appendix.
The anomaly ratio $\gamma$ for a dataset is mostly given between 0 and 0.2; $t_{e}-t_{s}$ is also determined by the dataset and generally ranges from 100 to 5,000 in the benchmark datasets.
\figurename~\ref{fig:f1score} depicts $\texttt{F1}_\texttt{PA}$~ varying with $\delta'$ under different $t_e-t_s$ when $\gamma$ is fixed to 0.05.
As shown in the figure, we can always obtain the $\texttt{F1}_\texttt{PA}$~close to 1 by changing $\delta'$, except for the case when the length of the anomaly segment is short.  


\subsection{Untrained model with comparably high \texttt{F1}}\label{sec:f1}
This section shows that the anomaly scores obtained from an untrained model are informative to a certain extent. 
A deep neural network is generally initialized with random weights drawn from a Gaussian distribution $\mathcal{N}\left(0, \sigma^2 \right)$, where $\sigma$ is often much smaller than 1.
Without training, the outputs of the model are close to zero because they also follow a zero-mean Gaussian distribution.
The anomaly score of a reconstruction-based or forecasting-based method is typically defined as the Euclidean distance between the input and output, which in the above case is proportional to the value of the input window:
\begin{equation}\label{eq:random_baseline}
    \begin{gathered}
        \anoscore = \left \| \bm{w}_{t} - \eta\; \right \|_2 \simeq \left \| \bm{w}_{t}\right \|_2,\\[3pt]
        \mathrm{where}\;\eta = f_\theta(\bm{w}_{t})\; \mathrm{and}\; \theta \sim \mathcal{N}(0, \sigma^2).
    \end{gathered}
\end{equation}
In the case of a point anomaly, the specific sensor values increase abruptly.  
This leads to a larger magnitude of $\left\| \bm{w}_t \right\|_2$ than normal windows, which is connected directly to a high $\anoscore$ for GT anomalies.
The experimental results in Section~\ref{section:results} reveal that \texttt{F1}~calculated from $\anoscore$ of Eq.~\ref{eq:random_baseline} is comparable to that of current TAD methods. 
It is also shown that \texttt{F1}~increases even more when the window size gets longer.  


\section{Towards a rigorous evaluation of TAD}\label{section:pak}
\subsection{New baseline for TAD}\label{sec:baseli}
For a classification task, the baseline accuracy is often defined as that of a random guess.
It can be said that there is an improvement only when the classification accuracy exceeds this baseline.
Similarly, TAD needs to be compared not only with the existing methods but also with the baseline detection performance.
Therefore, based on the findings of Section~\ref{sec:f1}, we suggest establishing a new baseline with the \texttt{F1}~measured from the prediction of a randomly initialized reconstruction model with simple architecture, such as an untrained autoencoder comprising a single-layer LSTM.
Alternatively, the anomaly score can be defined as the input itself, which is the extreme case of Eq.~\ref{eq:random_baseline} when the model consistently outputs zero regardless of the input. 
If the performance of the new TAD model does not exceed this baseline, the effectiveness of the model should be reexamined.


\subsection{New evaluation protocol \texttt{PA\%K}}\label{sec:pak}
In the previous section, we demonstrated that \texttt{PA}~has a great possibility of overestimating detection performance.
\texttt{F1}~without \texttt{PA}~can settle the overestimation immediately.
In this case, it is recommended to set a baseline as introduced in Section~\ref{sec:baseli}.  
However, depending on the test data distribution, \texttt{F1}~can unexpectedly underestimate the detection capability.
In fact, due to the incomplete test set labeling, some signals labeled as anomalies share more statistics with normal signals.
Even if anomalies are inserted intermittently over a period of time, $y_t=1$ for all $t$ in that period.

\begin{figure}[t!]
    \centering
    \includegraphics[width=\columnwidth]{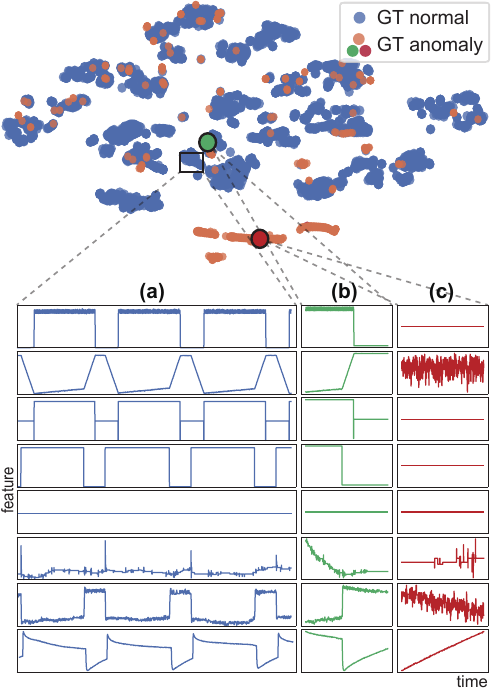}
    \caption{
    t-SNE of the input windows of the SWaT test dataset and visualization of corresponding signals. 
    Blue color indicates ground truth (GT) normal while orange, green and red color indicate GT anomaly.
    Even though (b) is GT anomaly, it shares patterns more with (a), GT normal signal, than (c) abnormal signal. 
    }
    \label{fig:tsne}
\end{figure}

We further investigated this problem using t-distributed stochastic neighbor embedding (t-SNE)~\cite{van2008visualizing}, as depicted in \figurename~\ref{fig:tsne}. 
The t-SNE is generated by the test dataset of secure water treatment (SWaT)~\cite{goh2016dataset}.
Blue and orange colors indicate the normal and abnormal samples, respectively.
The majority of the anomalies form a distinctive cluster located far from the normal data distribution.
However, some abnormal windows are closer to the normal data than anomalies.
The visualization of signals corresponding to the green and red points is depicted in (b) and (c), respectively.
Although both samples were annotated as GT anomalies, (b) shared more patterns with normal data of (a) than (c).
Concluding that the model's performance is deficient only because it cannot detect signals such as (b) can lead to an underestimation of the detection capability.

Therefore, we propose an alternative evaluation protocol \texttt{PA\%K}, which can mitigate the overestimation effect of $\texttt{F1}_\texttt{PA}$~and the possibility of underestimation of \texttt{F1}. 
Please note that it is not proposed to replace existing TAD metrics, but rather to be used along with them.
The idea of \texttt{PA\%K}~is to apply \texttt{PA}~to $S_m$ only if the ratio of the number of correctly detected anomalies in $S_m$ to its length exceeds the \texttt{PA\%K}~threshold \texttt{K}.
\texttt{PA\%K}~modifies Eq.~\ref{eq:palabeling} as follows: 
\begin{equation*}
    \hat{y}_{t} = \begin{cases}
    \multirow{2}{*}{1,} &\text{if}\;\anoscore > \delta \; \text{or}\\[2pt]
    &t \in S_m \; \text{and} \; \dfrac{\big| \{t' \mid t' \in S_m, \mathcal{A}\left(\bm{w}_{t'}\right) > \delta \}\big|}{\big|S_m\big|} > \texttt{K} \\ 
    0, &\text{otherwise}
    \end{cases}
\end{equation*}
where $\big| \cdot \big|$ denotes the size of $S_m$ (i.e., $t^m_e-t^m_s$) and \texttt{K} can be selected manually between 0 and 100 based on prior knowledge.
For example, if the test set labels are reliable, a larger \texttt{K} is allowable. 
If a user wants to remove the dependency on \texttt{K}, it is recommended to measure the area under the curve of $\texttt{F1}_\texttt{PA\%K}$~obtained by increasing \texttt{K} from 0 to 100.


\section{Experimental results}\label{section:results}
\subsection{Benchmark TAD datasets}

In this section, we introduce a list of the five most widely used TAD benchmark datasets as follows:

\noindent{\textbf{Secure water treatment (SWaT)}~\cite{goh2016dataset}}:
the SWaT dataset was collected over 11 days from a scaled-down water treatment testbed comprising 51 sensors~\cite{mathur2016swat}.
In the last 4 days, 41 anomalies were injected using diverse attack methods, while only normal data were generated during the first 7 days.

\noindent{\textbf{Water distribution testbed (WADI)}~\cite{wadi}}:
the WADI dataset was acquired from a reduced city water distribution system with 123 sensors and actuators operating for 16 days.
Only normal data were collected during the first 14 days, and the remaining two days contained anomalies. 
The test dataset had a total of 15 anomaly segments.

\noindent{\textbf{Server Machine Dataset (SMD)}~\cite{su2019robust}}:
the SMD dataset was collected from 28 server machines with 38 sensors for 10 days; only normal data appeared for the first 5 days, and anomalies were intermittently injected for the last 5 days. 
The results for the SMD dataset are the averaged values from 28 different models for each machine.

\noindent{\textbf{Mars Science Laboratory (MSL)}} and \textbf{Soil Moisture Active Passive (SMAP)}~\cite{hundman2018detecting}:
the MSL and SMAP dataset is a real-world dataset collected from a spacecraft of NASA. 
These are the anomaly data from an incident surprise anomaly (ISA) report for a spacecraft monitoring system. 
Unlike other datasets, unlabeled anomalies are contained in the training data, which makes training difficult.

\noindent{The statistics are summarized in Table~\ref{table:dataset_statistics}.}

\subsection{Evaluated methods}
Below, we present the descriptions of 7 representative TAD methods recently proposed and the 3 cases investigated in Section~\ref{section:method}.

\noindent{\textbf{USAD}~\cite{audibert2020usad}}
stands for unsupervised anomaly detection, which trains two autoencoders consisting of one shared encoder and two separate decoders, under a two-phase training scheme: an autoencoder training phase and an adversarial training phase. 

\noindent{\textbf{DAGMM}~\cite{zong2018deep}}
represents deep autoencoding Gaussian mixture model that adopts an autoencoder to yield a representation vector and feed it to a Gaussian mixture model. 
It uses the estimated sample energy as a reconstruction error; high energy indicates high abnormality. 

\noindent{\textbf{LSTM-VAE}~\cite{park2018multimodal}}
represents an LSTM-based variational autoencoder that adopts variational inference for reconstruction.

\noindent{\textbf{OmniAnomaly}~\cite{su2019robust}}
applied a VAE to model the time-series signal into a stochastic representation, which would predict an anomaly if the reconstruction likelihood of a given input is lower than a threshold value. 
It also defined the reconstruction probabilities of individual features as attribution scores and quantified their interpretability.

\noindent{\textbf{MSCRED}~\cite{zhang2019deep}}
represents a multi-scale convolutional recurrent encoder-decoder comprising convolutional LSTMs to reconstruct the input matrices that characterize multiple system levels, rather than the input itself. 


\noindent{\textbf{THOC}~\cite{shen2020timeseries}}
represents a temporal hierarchical one-class network, which is a multi-layer dilated recurrent neural network and a hierarchical deep support vector data description.


\noindent{\textbf{GDN}~\cite{deng2021graph}}
represents a graph deviation network that learns a sensor relationship graph to detect deviations of anomalies from the learned pattern.


\noindent{\textbf{Case 1. Random anomaly score}} corresponds to the case described in Section~\ref{sec:pa}.
The F1 score is measured with a randomly generated anomaly score drawn from a uniform distribution $\mathcal{U}$, i.e., $\anoscore \sim \mathcal{U}(0, 1)$.

\noindent{\textbf{Case 2. Input itself as an anomaly score}} denotes the case assuming $f_{\theta}(\bm{w}_t)=0$ regardless of $\bm{w}_t$. This is equal to an extreme case of Eq.~\ref{eq:random_baseline}.
Therefore, $\anoscore = \left \| \bm{w}_{t}\right \|_2$.

\noindent{\textbf{Case 3. Anomaly score from the randomized model}} corresponds to Eq.~\ref{eq:random_baseline}, where $\eta$ denotes a small output from a randomized model. 
The parameters were fixed after being initialized from a Gaussian distribution $\mathcal{N}(0, 0.02)$.
\begin{table}[t!]
\centering
\begin{tabular}{l|rrr}
\toprule
     Dataset &Train  &Test (anomaly\%) & $N$ \\
\midrule
\textbf{SWaT} & 495,000                 &449,919    (12.33\%)                  & 51   \\
\textbf{WADI} & 784,537                  & 172,801 (5.77\%)                    & 123  \\
\textbf{SMD}  & 25,300                   & 25,300 (4.21\%)                     & 38   \\
\textbf{MSL} &58,317 &73,729 (10.5\%) &55 \\
\textbf{SMAP} &135,183 &427,617 (12.8\%) &25 \\
\bottomrule
\end{tabular}
\caption{Statistics of benchmark TAD datasets. $N$ denotes the dimension of input features.}
\label{table:dataset_statistics}
\end{table}
\begin{table*}[t!]
\begin{adjustbox}{width=2.1\columnwidth,center}
\begin{tabular}{ccccccccccc}
\toprule
      & \multicolumn{2}{c}{\textbf{SWaT}} & \multicolumn{2}{c}{\textbf{WADI}} & \multicolumn{2}{c}{\textbf{MSL}} & \multicolumn{2}{c}{\textbf{SMAP}} & \multicolumn{2}{c}{\textbf{SMD}} \\ \cmidrule(lr){2-3} \cmidrule(lr){4-5} \cmidrule(lr){6-7} \cmidrule(lr){8-9} \cmidrule(lr){10-11}
      & $\texttt{F1}_\texttt{PA}$      & \texttt{F1}      & $\texttt{F1}_\texttt{PA}$      & \texttt{F1}     & $\texttt{F1}_\texttt{PA}$      & \texttt{F1}    & $\texttt{F1}_\texttt{PA}$      & \texttt{F1}      & $\texttt{F1}_\texttt{PA}$      & \texttt{F1}     \\
\midrule
USAD &0.846 \lo &\underline{0.791} \hi &0.429 \lo&0.232 \lo&\underline{0.927} \lo&0.211$^{\dagger}$ \lo &\underline{0.818} \lo&0.228$^{\dagger}$ \lo &\underline{0.938}\hi &0.426$^{\dagger}$ \lo\\
DAGMM &0.853 \lo &0.550 \lo&0.209 \lo&0.121 \lo&0.701 \lo&\rep{0.199} \lo &0.712 \lo&\rep{\textbf{0.333}} \hi &0.723 \lo  &\rep{0.238} \lo\\
LSTM-VAE &0.805 \lo &0.775 \lo&0.380 \lo&0.227 \lo&0.678 \lo&\rep{0.212} \lo &0.756 \lo&\rep{0.235} \hi &0.808 \hi &\rep{0.435} \lo \\
OmniAnomaly &0.866 \lo&0.782 \lo&0.417 \lo&0.223 \lo&0.899 \lo&\rep{0.207} \lo &0.805 \lo&\rep{0.227} \lo &\textbf{0.944} \hi &0.474$^{\dagger}$ \lo\\
MSCRED &0.868 \lo&\rep{0.662} \lo&0.346 \lo&\rep{0.087} \lo&0.775 \lo&\rep{0.199} \lo &\rep{0.942} \lo&\rep{0.232} \hi&  \rep{0.389} \lo&\rep{0.097} \lo\\ 
THOC &0.880 \lo&0.612$^{\dagger}$ \lo&0.506 \lo&\rep{0.130} \lo&0.891 \lo &0.190$^{\dagger}$ \lo &0.781$^{\dagger}$ \lo&0.240$^{\dagger}$ \hi& 0.541$^{\dagger}$ \lo & \rep{0.168} \lo  \\ 
GDN &\underline{0.935} \lo&\textbf{0.81} \hi&\underline{0.855} \lo&\textbf{0.57} \hi &0.903 \lo&0.217$^{\dagger}$ \lo &\rep{0.708} \lo &\underline{0.252}$^{\dagger}$ \hi &\rep{0.716} \lo & \textbf{0.529}$^{\dagger}$ \hi\\
\midrule
\textbf{Case 1} &\textbf{0.969} &0.216 &\textbf{0.965} &0.109 &\textbf{0.931} &0.190 &\textbf{0.961} &0.227 &0.804 &0.080 \\
\textbf{Case 2}   &0.873 &0.781 &0.694 &\underline{0.353} &0.812 & \underline{0.239} &0.675 &0.229 &0.896 &\underline{0.494} \\
\textbf{Case 3} &0.869 &0.789 &0.695 &0.331 &0.427 &\textbf{0.236} &0.699 &0.229 &0.893 &0.466  \\
\bottomrule
\end{tabular}
\end{adjustbox}
\caption{F1 score for various methods. $\dagger$ indicates the reproduced results. Bottom three rows represent the followings: \textbf{Case 1}. Random anomaly score, \textbf{Case 2}. Input itself as a anomaly score, \textbf{Case 3}. Anomaly score from a randomized model. Please refer to the manuscript for the detailed explanations. Bold and underlined cases indicate the best and the second best, respectively. \textcolor{green}{$\uparrow$} is marked in the following cases: (1) $\texttt{F1}_\texttt{PA}$~is higher than \textbf{Case 1}, (2) \texttt{F1}~is higher than \textbf{Case 2} or \textbf{3}.}
\label{table:random_comparsion}
\end{table*}

\subsection{Correlation between $\texttt{F1}_\texttt{PA}$~and \texttt{F1}}

\begin{figure}[t!]
    \centering
    \includegraphics[width=\columnwidth]{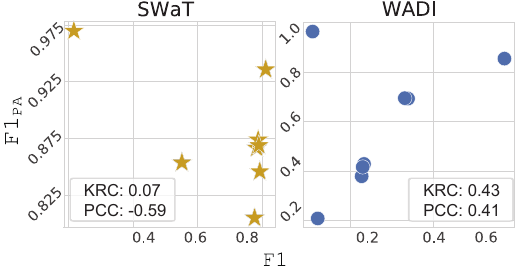}
    \caption{Correlation between $\texttt{F1}_\texttt{PA}$~and \texttt{F1}~of the existing methods on SWaT and WADI dataset. The Kendall rank correlation (KRC) and Pearson correlation coefficient (PCC) are indicated in the figure.}
    \label{fig:correlation}
\end{figure}

\texttt{F1}~is the most conservative indicator of detection performance. 
Therefore, if $\texttt{F1}_\texttt{PA}$~reliably represents the detection capability, it should have at least some correlation with \texttt{F1}.
\figurename~\ref{fig:correlation} plots $\texttt{F1}_\texttt{PA}$~and \texttt{F1}~for SWaT and WADI, as reported by the original studies on USAD, DAGMM, LSTM-VAE, OmniAnomaly, and GDN.
The figure also includes the results of \textbf{Case 1}--\textbf{3}. 
It is noteworthy that given that only a subset of the datasets and methods reported $\texttt{F1}_\texttt{PA}$~and \texttt{F1}~together, we plotted them only. 
For SWaT, the Pearson correlation coefficient (PCC) and Kendall rank correlation coefficient 
(KRC) were -0.59 and 0.07, respectively. 
For WADI, the PCC and KRC were 0.41 and 0.43, respectively.
However, these numbers are insufficient to assure the existence of correlation and confirm that comparing the superiority of the methods using only $\texttt{F1}_\texttt{PA}$~may have the risk of improper evaluation of the detection performance.


\subsection{Comparison results}\label{sec:window}
Here, we compare the results of the AD methods with \textbf{Case 1}--\textbf{3}.
It should be noted that the anomaly score is directly generated without model inference for \textbf{Case 1} and \textbf{2}.
For \textbf{Case 3}, we adopted the simplest encoder-decoder architecture with LSTM layers.
The window size $\tau$ for \textbf{Case 2} and \textbf{3} was set to 120.
For experiments that included randomness, such as \textbf{Case 1} and \textbf{3}, we repeated them with five different seeds and reported the average values. 
For the existing methods, we used the best numbers reported in the original papers and officially reproduced results~\cite{choi2021deep}; if there were no available scores, we reproduced them referring to the officially provided codes. 
Please note the we did not apply any preprocessing such as early time steps removal or downsampling.
The \texttt{F1}~for MSL, SMAP, and SMD have not been provided in previous papers; thus they are all reproduced.
It is worth noting that we searched for optimal hyperparameters within the suggested range in the papers and we did not apply down-sampling.
All thresholds were obtained from those that yielded the best score.
Further details of the implementation are provided in the Appendix.
The results are shown in Table~\ref{table:random_comparsion}. 
The reproduced results are marked as $\dagger$.
Bold and underlined numbers indicate the best and second-best results, respectively.
The up arrow (\textcolor{green}{$\uparrow$}) is displayed with the result for the following cases: (1) $\texttt{F1}_\texttt{PA}$~is higher than \textbf{Case 1}, (2) \texttt{F1}~is higher than \textbf{Case2} or \textbf{3}, whichever is greater.

Clearly, the randomly generated anomaly score (\textbf{Case 1}) is not able to detect anomalies because it reflects nothing about the abnormality in an input.
Correspondingly, \texttt{F1}~was quite low, which clearly revealed a deficient detection capability. 
However, when applying the \texttt{PA}~protocol, \textbf{Case 1} appears to yield the state-of-the-art $\texttt{F1}_\texttt{PA}$~far beyond the existing methods, except for SMD.
If the result is provided only with \texttt{PA}, as in the case of the MSL, SMAP, and SMD, distinguishing whether the method successfully detects anomalies or whether it merely outputs a random anomaly score irrelevant to the input is impossible.
In particular, \texttt{F1}~of the MSL and SMAP is quite low; this implies difficulty in modeling them, originating from the fact that they are both real-world datasets, and the training data contain anomalies.
However, $\texttt{F1}_\texttt{PA}$~appears considerably high, creating an illusion that the anomalies are being detected well for those datasets.

The \texttt{F1}~of \textbf{Case 1} of SMD is lower than that in other datasets, and there are previous methods surpassing it.
This may be attributed to the composition of the SMD test dataset.
According to Eqs.~\ref{eq:recall2} and \ref{eq:prec}, $\texttt{F1}_\texttt{PA}$~varies with three parameters: the ratio of anomalies in the test dataset $(\gamma)$, length of anomaly segments $(t_e-t_s)$, and TAD threshold $(\delta)$.
Unlike the other datasets, the anomaly ratio of SMD was quite low, as shown in Table~\ref{table:dataset_statistics}.
Moreover, the lengths of the anomaly segments are relatively short; the average length of 28 machines is 90, unlike other datasets ranging from hundreds to thousands.
This is similar to the lowest case in \figurename~\ref{fig:f1score}, which shows that the maximum achievable $\texttt{F1}_\texttt{PA}$~in this case is only approximately 0.8.
Therefore, we can conclude that the overestimation effect of \texttt{PA}~depends on the test dataset distribution, and its effect becomes less conspicuous with shorter anomaly segments. 

\begin{figure}[t!]
    \centering
    \includegraphics[width=\columnwidth]{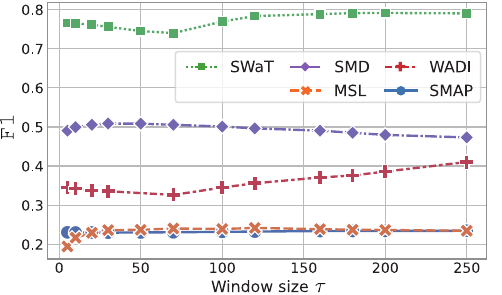}
    \caption{\texttt{F1}~for various window sizes $(\tau)$. As $\tau$ increases, \texttt{F1}~mostly increases after a short decrease. }
    \label{fig:windowsize}
\end{figure}

Across all datasets, the \texttt{F1}~for the existing methods is mostly inferior to \textbf{Case 2} and \textbf{3}, implying that the currently proposed methods may have obtained marginal or even no advancement against the baselines. 
Only the GDN consistently exceeded the baselines for all datasets.
The \texttt{F1}~of \textbf{Case 2} and \textbf{3} depend on the length of the input window.
With a longer window, the \texttt{F1}~baseline becomes even larger.
We experimented with various window lengths ranging from 1 to 250 in \textbf{Case 2} and depicted the results in Figure~\ref{fig:windowsize}. 
For SWaT, WADI, and SMAP, \texttt{F1}~begins to increase after a short decrease as $\tau$ increases.
This increase occurs because a longer window is more likely to contain more point anomalies, resulting in high anomaly score for the window.
If $\tau$ becomes too large, \texttt{F1}~saturates or degrades, possibly because the windows that used to contain only normal signals unexpectedly contain anomalies in it. 

\subsection{Effect of \texttt{PA\%K}~protocol}
\begin{figure}[t!]
    \centering
    \includegraphics[width=\columnwidth]{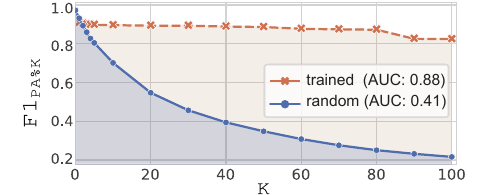}
    \caption{F1 score with $\texttt{PA\%K}$ with varying \texttt{K}. If $\texttt{K}=0$, it is equal to the $\texttt{F1}_\texttt{PA}$~and if $\texttt{K}=100$, it is equal to the \texttt{F1}.}
    \label{fig:k}
\end{figure}

To examine how \texttt{PA\%K}~alleviates the overestimation effect of \texttt{PA}~and underestimation tendency of \texttt{F1}, we observed $\texttt{F1}_\texttt{PA\%K}$~varying with different \texttt{PA\%K}~thresholds \texttt{K}.
\figurename~\ref{fig:k} shows the $\texttt{F1}_\texttt{PA\%K}$~for SWaT from \textbf{Case 1} and the fully trained encoder-decoder when \texttt{K} changes in increments of 10 from 0 to 100.
The $\texttt{F1}_\texttt{PA\%K}$~values of $\texttt{K}=0$ and $\texttt{K}=100$ are equal to the original $\texttt{F1}_\texttt{PA}$~and \texttt{F1}, respectively.
The $\texttt{F1}_\texttt{PA\%K}$~of a well-trained model is expected to show constant results regardless of the value of \texttt{K}.
Correspondingly, the $\texttt{F1}_\texttt{PA\%K}$ of the trained encoder-decoder (orange) shows consistently high $\texttt{F1}_\texttt{PA\%K}$.
In contrast, the $\texttt{F1}_\texttt{PA\%K}$ of \textbf{Case 1} (blue) rapidly decreased when \texttt{K} increased.
We also proposed measuring the area under the curve (AUC) to reduce the dependence on \texttt{K}. 
In this case, the AUC were 0.88 and 0.41 for the trained encoder-decoder and \textbf{Case 1}, respectively; this demonstrates that \texttt{PA\%K}~clearly distinguishes the former from the latter regardless of \texttt{K}.



\section{Discussion}\label{section:discussion}
Throughout this paper, we have demonstrated that the current evaluation of TAD has pitfalls in two respects: (1)~since \texttt{PA}~overestimate the detection performance, we cannot ensure that a method with higher $\texttt{F1}_\texttt{PA}$~has indeed a better detection capability; (2)~the results have been compared only with existing methods, not against the baseline.
A better anomaly detector can be developed when the current achievements are properly assessed. 
In this section, we suggest several directions for future TAD evaluations. 

The motivation of \texttt{PA}, i.e., the source of the first pitfall, originates from the incompleteness of the test dataset labeling process, as observed in Section~\ref{sec:pak}.
An exterminatory solution is to develop a new benchmark dataset annotated in a more fine-grained manner, so that the time-step-wise labels become reliable.
As it is often not feasible because fine-grained annotation requires tremendous resources, $\texttt{F1}_\texttt{PA\%K}$~can be a good alternative that can alleviate overestimation without any additional dataset modification. 
Please note that \texttt{PA\%K}~is an evaluation protocol that can be applied to various metrics other than the F1 score.
For the second issue, it is important to set a baseline as the performance of the untrained model as \textbf{Case 2} and \textbf{3} and measure the relative improvement against it.
The window size should be carefully determined by considering its effect on the baselines, as described in Section~\ref{sec:window}.

Furthermore, pre-defining the TAD threshold without any access to the test dataset is often impractical in the real world. 
Correspondingly, many AD methods in the vision field evaluate themselves using the area under the receiver operating characteristic (AUROC) curve~\cite{yi2020patch}.
In contrast, existing TAD methods set the threshold after investigating the test dataset or simply use the optimal threshold that yields the best \texttt{F1}.
Thus, the detection result depends significantly on threshold selection.
Additional metrics with the reduced dependency such as AUROC or area under precision-recall (AUPR) curve will help in rigorous evaluation.
Even in this case, the proposed baseline selection method is valid.
Since \texttt{PA\%K}~is a protocol, it also can be used to above metrics.

\section{Conclusion}\label{section:conclusion}
In this paper, we showed for the first time that applying \texttt{PA}~can severely overestimate a TAD model's capability, which may not reflect the true modeling performance.
We also proposed a new baseline for TAD and showed that only a few methods have achieved significant advancement in this regard.
To mitigate overestimation of \texttt{PA}, we proposed a new \texttt{PA\%K}~protocol that can be applied with existing metrics. 
Finally, we suggest several directions for rigorous evaluation of TAD methods, including baseline selection.
We expect that our research help clarify the potential of current TAD methods and lead the improvement of TAD in the future.
\section{Acknowledgement}
This work was supported by Institute of Information \& communications Technology Planning \& Evaluation (IITP) grant funded by the Korea government(MSIT) [No.2021-0-02068, Artificial Intelligence Innovation Hub, NO.2021-0-01343, Artificial Intelligence Graduate School Program (Seoul National University)], the National Research Foundation of Korea (NRF) grant funded by the Korea government (Ministry of Science and ICT) [2018R1A2B3001628, 2019R1G1A1003253], the Brain Korea 21 Plus Project in 2021, AIR Lab (AI Research Lab) in Hyundai Motor Company through HMC-SNU AI Consortium Fund, and Samsung Electronics. 

\bibliography{main.bib}

\setcounter{secnumdepth}{0}
\newpage
\section{A1~~~~Relaxation of assumptions}
In Section 3.2 of the manuscript, we demonstrated $\texttt{F1}_\texttt{PA}$ becomes close to 1 by restating recall and precision with conditional probability under two assumptions: (1) There exist only one anomaly segment in the test dataset (2) Anomaly scores are randomly drawn from a uniform distribution $\mathcal{U}(0, 1)$.
Here we relax the assumptions for generalization.

\subsection{A1.1~~~~Multiple anomaly segments}
We show that for the multiple anomaly segments, $\texttt{F1}_\texttt{PA}$ close to 1 is achievable without loss of generality. 
First, if there exists multiple anomaly segments in the dataset, i.e., $\boldsymbol{S}=\{S_1, ..., S_M\}$, then the recall $\mathrm{R}$ can be stated as follows:
\begin{equation}\label{eq:recall}
\begin{aligned}
    \mathrm{R}&=\mathbf{Pr}\left(\hat{y}_t=1 \mid y_t=1\right)\\
              &= 1 - \mathbf{Pr}(\hat{y}_t=0|t\in\boldsymbol{S})\\
              &= 1 - \frac{1}{\gamma} \cdot \mathbf{Pr}(\hat{y}_t=0, t \in \boldsymbol{S})\\
              &= 1 - \frac{1}{\gamma} \cdot \sum_{m=1}^M \mathbf{Pr}(\hat{y}_t=0, t\in S_m) \\
              &= 1 - \frac{1}{\gamma} \cdot \sum_{m=1}^M \mathbf{Pr}(\hat{y}_t=0) \cdot \mathbf{Pr}(t\in S_m) \\
              &= 1 - \frac{1}{\gamma} \cdot \sum_{m=1}^M \prod_{t'\; \in \;S_m} \mathbf{Pr}(\mathcal{A}\left( \bm{w}_{t'} \right) < \delta'))\cdot \mathbf{Pr}(t\in S_m) \\
              &= 1 - \frac{1}{\gamma} \cdot \sum_{m=1}^M \delta'^{(t^m_e-t^m_s)} \cdot \frac{t^m_e-t^m_s}{L}\\
              &= 1 - \sum_{m=1}^M \delta'^{(t^m_e-t^m_s)} \cdot \frac{(t^m_e-t^m_s)}{\sum_{m=1}^M t^m_e-t^m_s},\\
\end{aligned}
\end{equation}
\noindent{where} $L$ denotes the length of the input signal, i.e., $L \cdot \gamma = \sum_{m=1}^M t^m_e-t^m_s$.
Then, the precision $\mathrm{P}$ is written as below:
\begin{equation}\label{eq:precision}
    \begin{aligned}
        \mathrm{P} &= \mathrm{R} \cdot \frac{\mathbf{Pr}(y_t=1)}{\mathbf{Pr}(\hat{y}_t=1)}\\
                   &= \frac{\gamma\cdot\mathrm{R}}{\mathbf{Pr}(\hat{y}_t=1, y_t=0)+\mathbf{Pr}(\hat{y}_t=1, y_t=1)}\\
                   &= \frac{\gamma\cdot\mathrm{R}}{(1-\delta')(1-\frac{\sum_{m=1}^M t^m_e-t^m_s}{L}) + \gamma \cdot \mathrm{R}}.
    \end{aligned}
\end{equation}

\begin{figure}[t!]
    \centering
    \includegraphics[width=\columnwidth]{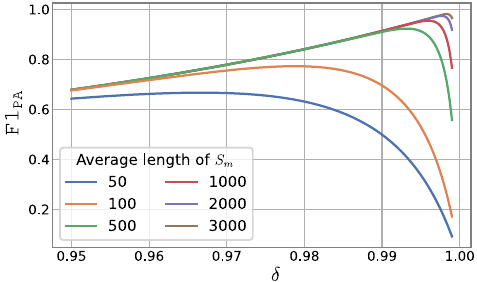}
    \caption{
    $\texttt{F1}_\texttt{PA}$ for $\mathcal{A}(\boldsymbol{w}_t)\sim \mathcal{U}(0, 1)$, $\gamma=0.05, M=6$
    }
    \label{fig:appendix_threshold}
\end{figure}

\noindent Figure~\ref{fig:appendix_threshold} depicts the $\texttt{F1}_\texttt{PA}$ for various thresholds, when $\gamma=0.05$ and $M=6$.
If the average length of $S_m$ is sufficiently large, which is typical for the majority of benchmark datasets, $\texttt{F1}_\texttt{PA}$ achieves close to 1 depending on $\delta'$.

\subsection{A1.2~~~~Anomaly scores drawn from a Gaussian distribution}
\begin{figure}[t!]
    \centering
    \includegraphics[width=\columnwidth]{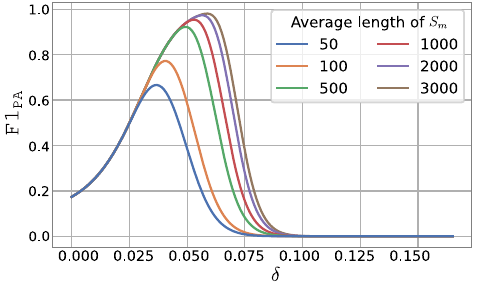}
    \caption{
    $\texttt{F1}_\texttt{PA}$ for $\mathcal{A}(\boldsymbol{w}_t)\sim \mathcal{N}(0, 0.02)$, $\gamma=0.05, M=6$
    }
    \label{fig:appendix_threshold2}
\end{figure}

Here, we extend the above $\texttt{F1}_\texttt{PA}$ for multiple anomaly segments case to the anomaly scores drawn from drawn from a Gaussian distribution, i.e., $\mathcal{A}(\boldsymbol{w}_t) \sim \mathcal{N}(0, \sigma$), where $0< \sigma << 1$.
Then, $\mathbf{Pr}(\mathcal{A}(\boldsymbol{w}_t)<\delta')$ of Eq.~\ref{eq:recall} changes to $\int_{\delta'}^{\infty} \mathcal{N}(0, \sigma)$.
In this case, $\texttt{F1}_\texttt{PA}$ also achieves close to 1 if the average length of window, as shown in Figure.~\ref{fig:appendix_threshold2}, where $\sigma$ is 0.02.
Anomaly scores generated from a Gaussian distribution corresponds to the untrained model initialized with the same distribution.

\section{A2~~~~Related works}
\textbf{Reconstruction-based AD} method trains a model to minimize the distance between a normal input and its reconstruction.
Anomalous input at the test time results in large distance as it is difficult to reconstruct.
The distance, or reconstruction error, serves as an anomaly score.
Various network architectures have been adopted for reconstruction, from an autoencoder~\cite{malhotra2016lstm} to a generative adversarial network (GAN)~\cite{li2019mad, zhou2019beatgan}. 
The distance metrics also vary from the Euclidean distance~\cite{audibert2020usad} to likelihood of reconstruction~\cite{su2019robust}.

\textbf{Forecasting-based AD} method is similar to the reconstruction-based methods, except that it predicts a signal for the future time steps.
The distance between the predicted and ground truth signal is considered an anomaly score. 
\citeauthor{hundman2018detecting} adopted a long short-term memory (LSTM)~\cite{hochreiter1997long} to forecast the current time-step-ahead input.  
Another variant of the recurrent model, the gated recurrent unit~\cite{chung2014empirical}, was applied to the forecasting-based TAD~\cite{wu2020developing}.


\textbf{Others} include various attempts to model the normal data distribution.
One-class classification-based approaches~\cite{ma2003timeseries, shen2020timeseries} measured the similarity between the hidden representations of the normal and abnormal signals.
Real-time anomaly detection in multivariate time-series (RADM)~\cite{ding2018radm} applied hierarchical temporal memory and a Bayesian network to model normal data distribution. 
Recently, as graph neural networks (GNNs) have shown very good performance in modeling temporal dependency, their use in TAD is growing rapidly~\cite{chen2021learning, deng2021graph}.

\section{A3~~~~Experimental details}
\subsection{A3.1~~~~Experimental environment}
All experiments, including the reimplementations, were conducted with the following environment:
\begin{itemize}
    \item \textbf{CPU}: Intel(R) Xeon(R) Gold 6242R
    \item \textbf{GPU}: NVIDIA GeForce RTX 2080Ti (MSCRED, THOC, GDN)
    Quadro RTX 8000 with 48GB VRAM (remainder)
    \item \textbf{Library}: PyTorch~\cite{NEURIPS2019_9015} and Tensorflow~\cite{tensorflow2015-whitepaper}
    \item \textbf{CUDA version}: 11.0
\end{itemize}

\subsection{A3.2~~~~Details of reimplementation}
To reproduce the results of existing methods, we referred to the public codes published by the authors.
Although some methods, including USAD~\cite{audibert2020usad} and GDN~\cite{deng2021graph}, state that they applied down-sampling in the original paper, we did not apply it in the main manuscript because it potentially modifies the test dataset distribution.
Moreover, as previously reported in OmniAnomaly~\cite{su2019robust}, applying down-sampling at various sampling rates did not make a significant difference in results.
We used default hyper-parameters provided by the public code except for some hyper-parameters that depend on the dataset and they are listed below.
For all methods and datasets presented in the manuscript, we experimented with three different window sizes in a list incremented by 10 from 10 to 100.
Below are the list of hyper-parameters that we searched for:
\begin{itemize}
    \item \textbf{USAD}
        \begin{itemize}
            \item context vector size: [10, 50, 100]
        \end{itemize}
    \item \textbf{DAGMM}~\cite{zong2018deep}
    \begin{itemize}
        \item unit sizes of hidden layers of compression network: [32, 64, 80] 
        \item unit sizes of hidden layers of estimation network: [16, 32, 48]
    \end{itemize}
    \item \textbf{LSTM-VAE}
    \begin{itemize}
        \item intermediate dimension: fixed to 200 
        \item dimension of latent vector: [15, 30, 50]
    \end{itemize}
    \item \textbf{OmniAnomaly}
    \begin{itemize}
        \item dimension of latent vector: [3, 5, 10]
        \item hidden unit of recurrent neural network (RNN): [300, 500, 700]
    \end{itemize}
    \item \textbf{THOC}~\cite{shen2020timeseries}
    \begin{itemize}
        \item warmup length for dilated RNN test data: [500, 1000]
        \item prediction loss lambda: [0.01, 0.1, 1, 10, 100]
        \item orthogonality loss lambda: [0.01, 0.1, 1, 10, 100]
        \item step sizes for 3-layere dilated RNN: [\{1, 2, 4\},\{1, 4, 8\},\{1, 4, 12\}, \{1, 4, 16\}]
        \item hidden unit dimension: [32, 64, 84, 128]
        \item number of clusters for the 3-layer dilated RNN: [\{6, 6, 6\}, \{12,6,1\}, \{12,6,4\},\{18,6,1\}\{18, 12, 4\}\{18, 12, 6\}\{32, 12, 6\}]
    \end{itemize}
    \item \textbf{GDN}:
    \begin{itemize}
    \item embedding dimension: [64, 128, 256]
    \item number of output layers: [1, 2, 4]
    \item number of adjacent attributes: [5, 15]
    \item intermediate dimension of output layer: [64, 128, 256]
    \end{itemize}
    
\end{itemize}

\subsection{A3.3~~~~Architecture of the untrained model}
In experimental results, we introduced a simple untrained encoder-decoder for implementation of \textbf{Case 3. Anomaly score from the randomized model}.
Below is the description of its architecture:
\begin{itemize}
    \item \textbf{Encoder}: Single-layer LSTM
    \item \textbf{Decoder}: Single-layer LSTM
    \item \textbf{Hidden unit dimension}: 25
    \item \textbf{Context dimension}: 25
\end{itemize}
\noindent{We} note that the model was not trained after initialization.

\end{document}